\documentclass[11pt]{article}
\usepackage{irvlreport}

\usepackage{amsmath}      
\usepackage{booktabs}     
\usepackage{graphicx}     
\usepackage{url}          
\usepackage{listings}     
\usepackage{natbib}       
\usepackage{subcaption}   
\usepackage{placeins}
\usepackage{enumitem}
\usepackage{float}

\usepackage{algorithm}
\usepackage{algpseudocode}

\usepackage{array,multirow}

\newcommand{\z}{\textcolor{black!35}{--}}
\newcommand{\sbar}[1]{%
  \rlap{\textcolor{black!10}{\rule[-0.25ex]{34pt}{6.5pt}}}%
  \ifnum#1>0\rlap{\textcolor{black!68}{\rule[-0.25ex]{\dimexpr#1\dimexpr3.4pt\relax\relax}{6.5pt}}}\fi
  \hspace*{34pt}}

\newcommand{\permodelcap}{\captionsetup{position=above,skip=5pt,%
  justification=raggedright,singlelinecheck=false}}
  
\graphicspath{{figures/}{./}}

\irvlsettitle{From Reasoning to Action: Benchmarking Frontier AI Models on LIBERO-10 Robot Manipulation Tasks}
\irvlsetshorttitle{LIBERO-10 Benchmark}
\irvlsetauthors{May Espinola\textsuperscript{*}, Qifan Zhang\textsuperscript{*}, Yu Xiang\\[3pt]
\normalsize\textsuperscript{*}Equal contribution}
\irvlsetdate{October 2026}
\irvlsetversion{Version 1.0}
\irvlsetlogo{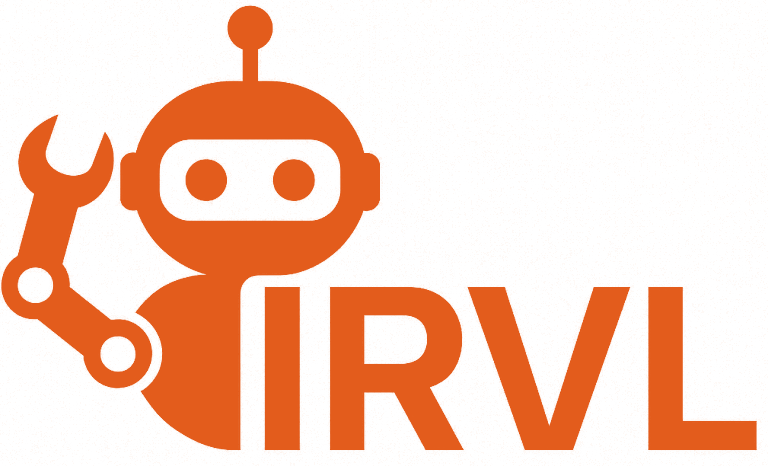}
\irvlsetprojecturl{https://irvlutd.github.io/LLM-LIBERO/}
\irvlsetcodeurl{https://github.com/IRVLUTD/LLM_LIBERO}

\providecommand{\taskpanel}[3]{%
  \begin{subfigure}[t]{#1\textwidth}
    \includegraphics[width=0.495\linewidth]{agentview/task_#2_init00.png}\hfill%
    \includegraphics[width=0.495\linewidth]{wrist/task_#2_init00.png}
    \caption{#3}
  \end{subfigure}}
\begin{document}
\irvlmaketitle

\begin{irvlabstract}
Recent advances in frontier AI models have demonstrated remarkable capabilities in visual understanding, reasoning, planning, and tool use, raising the possibility of using general-purpose AI models for robot manipulation without task-specific training. However, their manipulation capabilities have not been systematically evaluated on standardized robotics benchmarks. In this report, we present a unified closed-loop manipulation framework for evaluating frontier AI models on the LIBERO-10 benchmark. Our framework enables models to control a robot directly through visual observations, robot proprioception, and structured tool calls, primarily by specifying target end-effector poses and gripper commands, without task-specific fine-tuning or pretrained manipulation skills. We evaluate six frontier AI models, including GPT-6 Astra, GPT-6.1 Sol, GPT-6 Luna, Claude Opus 5, Claude Opus 5.5, and Gemini-4 Argon, across three reasoning effort levels, resulting in 18 configurations and 1,800 evaluation episodes. Our results show that GPT-6 Astra achieves the highest success rate of 44\%, followed by Gemini-4 Argon at 43\%. Increasing reasoning effort generally improves manipulation performance, but the benefits vary across models and come at additional inference cost. We also observe substantial differences across tasks, with models performing relatively well on object placement tasks but struggling with precise spatial arrangements and complex multi-step interactions. These findings demonstrate both the potential and limitations of frontier AI models for robot manipulation, highlighting the challenges of translating reasoning capabilities into reliable physical actions. We release our evaluation framework and experimental results to support reproducible benchmarking and future research on reasoning-based robot manipulation.
\end{irvlabstract}


\section{Introduction}

Recent advances in frontier artificial intelligence (AI) models have demonstrated remarkable capabilities in visual understanding, reasoning, planning, and tool use~\cite{arena2026leaderboard}. In particular, recent demonstrations of GPT-6 Astra have shown impressive robot manipulation behaviors, including interpreting visual observations, generating robot actions, and correcting mistakes through closed-loop interaction~\cite{zhang2026unexpected,su2026astra,team2026systematically}. NVIDIA's \emph{Astranomy} report~\cite{yang2026astranomy} further demonstrates the potential of frontier AI models for robot manipulation through different control configurations, including direct action generation, agentic tool use, and supervision of pretrained robot policies. These results suggest that general-purpose AI models may be capable of performing robot manipulation without task-specific policy training.

Despite these promising demonstrations, an important question remains: \textbf{How well do frontier AI models perform on established robot manipulation benchmarks?} Existing evaluations often use different tasks, robot platforms, and control interfaces, making systematic comparisons difficult. Connecting frontier AI models to standardized benchmarks is essential for understanding their capabilities, identifying limitations, and comparing their performance with conventional robot learning approaches.

Motivated by this question, we develop a unified manipulation pipeline to evaluate frontier AI models on LIBERO-10 (also known as LIBERO-Long)~\cite{liu2023libero}, a subset of LIBERO-100 containing ten held-out, long-horizon manipulation tasks. LIBERO-10 has been widely adopted for evaluating robot learning methods, including generalist robot policies and vision-language-action models such as Octo~\cite{team2024octo}, OpenVLA~\cite{kim2024openvla}, $\pi_{0.5}$~\cite{intelligence2025pi_}, UniVLA~\cite{bu2025univla} and UniAct~\cite{zheng2025universal}. Built upon the open-source Inspect Robots framework~\cite{inspect-robots}, our pipeline enables frontier AI models to interact with the simulation environment through visual observations and structured tool calls, allowing closed-loop manipulation without task-specific training or fine-tuning. We evaluate multiple frontier AI models, including GPT and Claude models, under a common protocol, exploring different reasoning modes to investigate the impact of reasoning effort on manipulation performance. Beyond task success rates, we analyze API calls, inference latency, token usage, and monetary cost to understand the tradeoffs between reasoning capabilities, manipulation performance, and efficiency.

Our work provides a standardized evaluation of frontier AI models on robot manipulation tasks, offering insights into their capabilities and limitations. By releasing our code and evaluation results, we provide reproducible, ready-to-use baselines for future research, reducing the need to repeatedly evaluate these computationally expensive models and lowering evaluation costs.

\section{Closed-Loop Manipulation with Frontier AI Models}
\label{sec:method}

\begin{figure}[!ht]
\vspace{-10pt}
\centering
\includegraphics[width=\textwidth]{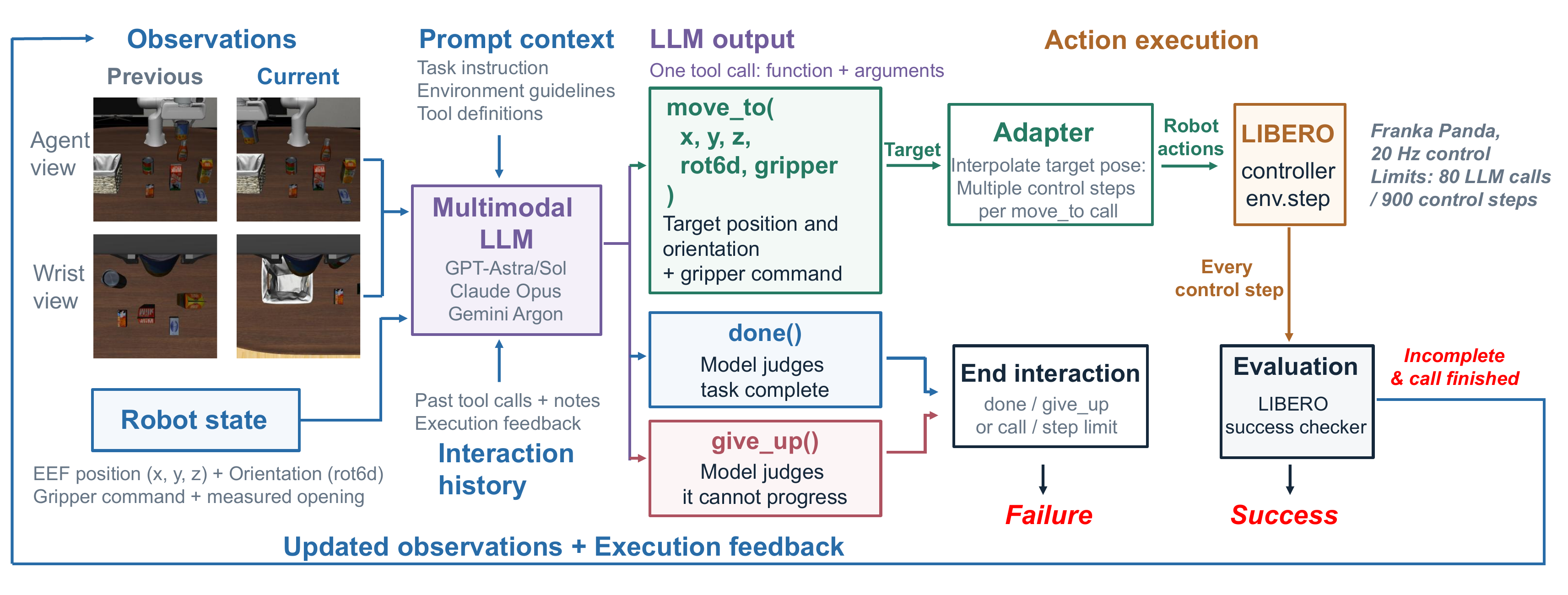}
\caption{\textbf{Overview of our closed-loop manipulation framework using frontier AI models.} }
\label{fig:framework}
\end{figure}

We evaluate frontier AI models, including GPT-6, Claude Opus, and Gemini, directly as closed-loop robot manipulation policies, without robot-specific training, demonstrations, or pretrained skill libraries. Unlike approaches that use AI models to generate robot code or orchestrate predefined skills~\cite{xiao2026enpire,lu2026aspire,chen2026gap,ora2026}, our pipeline focuses on evaluating the models' intrinsic visual understanding, spatial reasoning, and action planning capabilities. Given raw camera observations, the models must identify objects, estimate their 3D positions, and plan robot motions through a minimal action interface, without relying on external perception, planning, or manipulation skills. This design allows us to assess how effectively frontier AI models can translate their own reasoning into robot actions.

Figure~\ref{fig:framework} illustrates our closed-loop manipulation pipeline. At each interaction step, the model receives a fixed system prompt, the task instruction, the interaction history, and the latest observation, including camera images and robot state information (Sec.~\ref{sec:observations}). Based on these inputs, the model generates a single tool call: \texttt{move\_to} specifies a target end-effector pose and gripper command, while \texttt{done} and \texttt{give\_up} terminate the episode (Sec.~\ref{sec:actions}). For each \texttt{move\_to} call, an action adapter translates the target into a sequence of low-level robot commands, which are executed step by step in the LIBERO simulator. Task success is checked after every control step. If the task remains incomplete after the action sequence finishes, updated observations and execution feedback are returned to the model for the next decision (Sec.~\ref{sec:evaluation}). This process continues until the task succeeds, the model terminates the episode, or the evaluation limits are reached. All experiments are conducted in a \emph{zero-shot setting}, using a task-agnostic system prompt without task-specific hints or demonstrations.

\subsection{Observations and Prompt Context}
\label{sec:observations}

\paragraph{Observations.}
Each observation contains two $256\times256$ RGB images, from the fixed third-person (agentview) camera and the wrist camera, together with the robot state. The robot state has 12 values: the end-effector position $(x, y, z)$ in meters in the world frame, its orientation in the 6D rotation representation (the first two columns of the rotation matrix), the current gripper command, and the two measured finger positions. The gripper command only records whether the model last asked the gripper to open or close; the measured finger positions show where the fingers actually stopped, so the model can tell whether it closed on an object or on nothing. Each observation is sent as one message: the robot state as text, followed by the two labeled camera images. The model never sees object poses or the success signal.

\paragraph{System prompt.}
The system prompt, reproduced verbatim in Figure~\ref{fig:prompt}, has two parts. The first part is a generic agent instruction that is independent of the robot and the task: it explains the tool-call interface, requires exactly one tool call per turn with a short note on every motion, tells the model when to call \texttt{done} or \texttt{give\_up}, and states the budget of 80 LLM calls. The second part, appended under \emph{Embodiment notes}, contains the environment guidelines that we wrote for LIBERO. They describe the robot, the coordinate frame, the rotation representation, the gripper behavior, and the available sensing, followed by eight general manipulation rules. The same prompt is used for all ten tasks and all models; only the embodiment name changes with the task. The task instruction is not part of the system prompt. It is LIBERO's original natural-language task description, which the model receives as the first user message and again at the top of every observation message. The instructions of all ten tasks are listed in Figure~\ref{fig:taskgallery}.

\begin{figure}[t]
\centering
\begin{lstlisting}[basicstyle=\ttfamily\scriptsize, breaklines=true, breakindent=0pt, columns=fullflexible, frame=single, framesep=4pt]
You are controlling a real robot embodiment named 'libero_10_task00' through tool calls. Each observation message gives you the current proprioceptive state and camera images. Work toward the user's goal in small, deliberate motions; re-check the observation after every motion. Every move tool call must include a `note`: in one or two sentences, say what you observe in the current observation and why you chose this motion. The user is watching these notes to see what you see and what you decide, so write them for a human reader. Safety approvers clamp out-of-bounds and too-fast actions below you. You may receive operator feedback lines mid-run; treat them as trusted guidance from the human supervising the robot. Respond with exactly one tool call per turn. When the goal is achieved call done; if it cannot be achieved call give_up. Note what you are learning about this rig and task as you go: done and give_up will ask what you wish you had known from the start. You have a budget of 80 LLM calls for the whole trial.

Embodiment notes:
Single Panda arm. move_to uses world-frame x,y,z in meters and rot6d orientation (first two rotation-matrix columns). z points up. Gripper 0=open, 1=close. The gripper is open/close only: any value above 0.5 keeps closing and any value below 0.5 keeps opening, so intermediate values do not give a partial opening. The gripper entry of eef_state is the current gripper command, not the finger position: it reads 1 while closed on an object even though the fingers are stopped apart. Measured finger positions are in gripper_qpos (meters, about +/-0.04 each when fully open, near 0 when closed on nothing); fingers stopped clearly apart after closing indicate something is between them. Omitting gripper from a move_to call keeps the current command. Keep orientation unchanged unless necessary. Images are upright agentview and wrist camera views. Only robot proprioception and RGB are available; no object coordinates. Intermediate target arrival must be checked from the next observation. Verify the object actually followed the lift in the next observation before assuming the grasp is secure.

General manipulation guidance:
1. Avoid collisions: before a lateral move that could intersect nearby objects or furniture, lift to a visibly safe clearance. Do not sweep through or push objects unless the task requires it. Use large moves in free space and small moves only in the last few centimeters near an object.
2. Identify the target object carefully; similar-looking distractors may be present. Cross-check both camera views.
3. A closed gripper does not mean a successful grasp. After closing, lift 3-5 cm and confirm in both cameras that the object rose with the gripper.
4. Handle objects the way a person would: grasp a stable, graspable part without pushing or tipping the object, keep it in the same orientation (usually upright) from grasp to placement, and lower it close to the support surface before releasing.
5. If the same approach fails twice, change it (grasp point, approach direction, or a small adjustment); do not repeat the same motion.
6. After each move, confirm from the next observation that it succeeded before starting the next step; do not assume success.
7. Prefer fewer, purposeful moves over many tiny corrections; your call budget is limited.
8. Declare success only when the latest observation clearly shows that the requested final state has been achieved.
\end{lstlisting}
\vspace{-3pt}
\caption{\textbf{Exact system prompt}. The first paragraph is the generic agent instruction; the text after \emph{Embodiment notes} is our environment guidelines. Only the embodiment name differs across tasks.}

\label{fig:prompt}
\end{figure}

\paragraph{Interaction history.}
Each LLM call is stateless, so every call resends the fixed context (the system prompt and the task instruction), followed by the interaction history accumulated so far. Text and images in this history are treated differently. The text history is complete: it contains the text part of every observation since the start of the episode (the task instruction and the 12 robot-state values), the model's own tool calls and notes, and the execution feedback of each call. Images, in contrast, are kept only for the two most recent observations (the \emph{previous} and \emph{current} observations in Figure~\ref{fig:framework}), i.e., at most four images per call; the images of all earlier observations are replaced by a short text placeholder. Because each call adds one observation, the length of the history is bounded by the budget of 80 LLM calls.

\begin{figure}[t]
\centering
\begin{lstlisting}[basicstyle=\ttfamily\scriptsize, breaklines=true, breakatwhitespace=true, breakindent=0pt, columns=fullflexible, frame=single, framesep=4pt]
move_to(targets, note)
  Move to absolute Cartesian end-effector targets (meters for positions, radians for rotations, per the dimension labels). The motion is a straight line interpolated at a fixed safe speed and the result reports its step count. Unnamed dimensions hold their current value. Coordinates are absolute in the embodiment's declared frame; on multi-arm embodiments each arm uses its own base frame and axes may differ between arms depending on mounting. Rotation dimensions are absolute targets measured relative to the trial's start orientation (0 means the start orientation) and interpolate linearly without wrapping, so prefer intermediate values for large rotations. Per-dimension bounds: x: [-0.6, 0.6], y: [-0.6, 0.6], z: [0.42, 1.6], r00: [-1, 1], r10: [-1, 1], r20: [-1, 1], r01: [-1, 1], r11: [-1, 1], r21: [-1, 1], gripper: [0, 1].
  - targets (required): Map of dimension name to value. Valid names: x, y, z, r00, r10, r20, r01, r11, r21, gripper
  - note (required): What you observe right now in the observation (images, if any, and state), and why you chose this motion. The user reads these notes live and in the saved transcript to follow what you see and what you decide. Write for them, in one or two plain sentences.

done(summary, hindsight)
  Declare the task finished. The trial ends; a scorer judges success.
  - summary (required)
  - hindsight (required): see below

give_up(reason, hindsight)
  Stop trying; the task cannot be completed. The trial ends.
  - reason (required)
  - hindsight (required): What do you know now that you wish you had known at the start of this episode? Concrete, transferable facts about this rig, task, or embodiment (camera mounting and extrinsics, table and base geometry, gripper axis and offsets, controller behavior, metric scale), written as advice to a future agent attempting the same task. Say 'none' if nothing qualifies.
\end{lstlisting}
\vspace{-3pt}
\caption{\textbf{Tool definitions given to the model.} The descriptions are reproduced verbatim; they are sent as JSON function schemas with every request.}
\label{fig:tooldefs}
\end{figure}

\begin{figure}[h]
\centering
\includegraphics[width=0.85\textwidth]{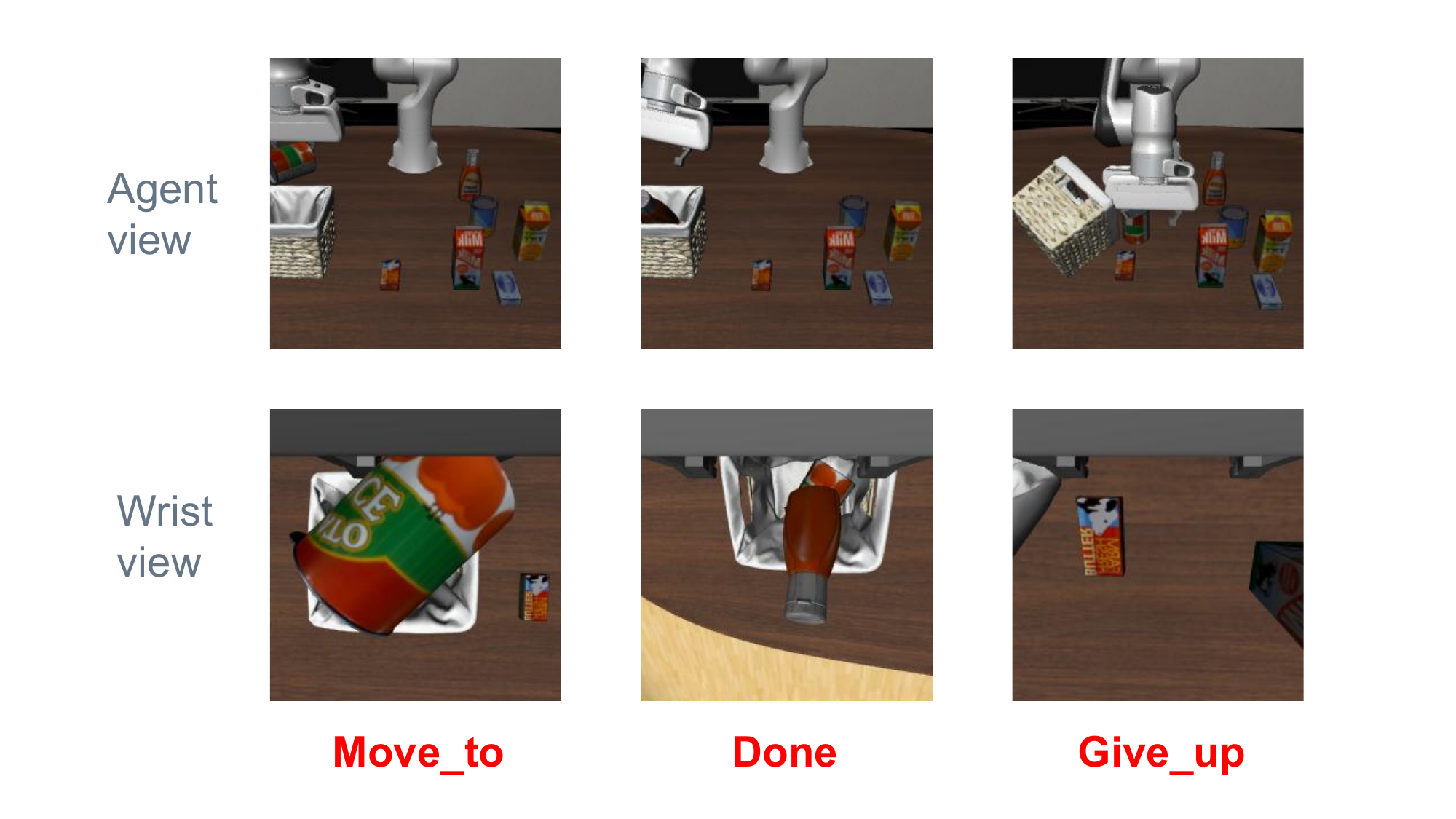}
\caption{\textbf{Examples of the three tool calls} in task~0 (``put both the alphabet soup and the tomato sauce in the basket''), taken from Claude Opus~5.5 (low effort). Each column shows the agentview (top) and wrist (bottom) images on which the model issued the call.
\textbf{\texttt{move\_to}:} holding a can over the basket, the model calls \texttt{move\_to} to move the end effector to an absolute target pose and lower the can into the basket.
\textbf{\texttt{done}:} the model believes the task is complete, but it has placed the wrong objects in the basket (the tomato sauce can and a ketchup bottle, while the alphabet soup remains on the table), so the episode fails.
\textbf{\texttt{give\_up}:} the arm knocks the basket onto its side; the model judges that the task can no longer be completed and gives up, so the episode fails.}
\label{fig:tools}
\end{figure}

\subsection{Action Interface and Execution}
\label{sec:actions}

\paragraph{Tools.}
The model acts through three tools, one tool call per turn. Their full definitions are given in Figure~\ref{fig:tooldefs}, and Figure~\ref{fig:tools} shows one example of each. The tool calls are generated through the native function-calling interface of each model's API: every request includes the name, description, and argument format of each tool, and the model replies with a structured call that names one tool and gives its arguments in JSON, which the framework then executes. For instance, in one task-0 episode, Claude Opus~5.5 (high effort) returned the following \texttt{move\_to} call, which sets all ten dimensions to rotate the gripper by $90^\circ$ about the vertical axis while moving over a fallen can:
\begin{lstlisting}[basicstyle=\ttfamily\footnotesize, columns=fullflexible, frame=single, framesep=3pt, xleftmargin=3pt, xrightmargin=3pt, aboveskip=6pt, belowskip=4pt]
{"type": "tool_use",
 "name": "move_to",
 "input": {
   "targets": {"x": -0.08, "y": 0.115, "z": 0.55,
               "r00": -1.0, "r10": 0.0, "r20": 0.0,
               "r01": 0.0, "r11": 1.0, "r21": 0.0,
               "gripper": 0},
   "note": "The alphabet soup is lying on its side on the table between
            the basket and the milk carton, with its axis along world y.
            I'm finishing the 90 deg yaw so the fingers close along x,
            across the can, and moving over it at about x=-0.08, y=0.115
            while lowering to check alignment."
 }}
\end{lstlisting}

\begin{itemize}
    \item \texttt{move\_to(targets, note)} moves the end effector to an absolute target pose (Figure~\ref{fig:tools}, left). In each call, the model outputs two arguments. \texttt{targets} gives target values for any subset of the ten action dimensions: the position $(x, y, z)$ in meters in the world frame, the six entries of the 6D rotation representation, and the gripper command. \texttt{note} is a short explanation of what the model observes and why it chose the motion. The tool description tells the model that the end effector moves along a straight line at a fixed safe speed, that unspecified dimensions keep their current value, and that each dimension is bounded: $x, y \in [-0.6, 0.6]$\,m, $z \in [0.42, 1.6]$\,m, rotation entries in $[-1, 1]$, and the gripper command in $[0, 1]$.

    \item \texttt{done(summary, hindsight)} lets the model end the episode when it believes the task is complete (Figure~\ref{fig:tools}, middle). No action is executed, and the episode ends immediately. A \texttt{done} call is never scored as a success, for the following reason. LIBERO's success checker runs after every control step, and as soon as the task is complete it ends the episode as a success (Sec.~\ref{sec:evaluation}). The model is therefore asked for its next call only when the checker has just reported that the task is not yet complete. Since \texttt{done} does not move the robot, the task is still incomplete when the model calls it, so the episode counts as a failure. In other words, a \texttt{done} call means that the model believes it has finished when it has not.

    We still provide \texttt{done} because the model never sees the success checker and must decide by itself when the task is finished. Without \texttt{done}, a model that believes it has finished would have no way to stop and would keep acting. The frequency of \texttt{done} calls therefore indicates how often a model mistakes an unfinished task for a finished one.

    \item \texttt{give\_up(reason, hindsight)} declares that the model believes the task cannot be completed, for example after repeated failed grasps or when an object or the container has been knocked out of reach (Figure~\ref{fig:tools}, right); \texttt{reason} explains why. As with \texttt{done}, no action is executed and the episode ends immediately as a failure. The framework also issues \texttt{give\_up} on the model's behalf when the budget of 80 LLM calls is exhausted.
\end{itemize}

\paragraph{Hindsight.}
Both \texttt{done} and \texttt{give\_up} also require a \texttt{hindsight} field, in which the model states what it wishes it had known at the start of the episode, such as camera placement, controller behavior, or metric scale, written as advice for a future attempt. These notes are logged for analysis but are never fed back to the model, so every episode remains zero-shot.

\paragraph{Execution of \texttt{move\_to}.}
The adapter (Figure~\ref{fig:framework}) interpolates a straight-line trajectory from the current end-effector pose to the target at a fixed safe speed, producing a sequence of absolute end-effector actions at 20\,Hz (about eleven actions per call on average). The number of interpolated steps is returned to the model as the tool result and kept in the interaction history. The actions are then executed one at a time: each 10-D action is converted to LIBERO's 7-D absolute pose command by orthonormalizing the 6D rotation and converting it to axis-angle form, and LIBERO's operational space controller (OSC) drives the Franka Panda arm for one control step.

\begin{algorithm}[t]
\caption{One closed-loop episode. $o$: observation (agentview and wrist images plus the 12-value robot state); $H$: prompt context and interaction history sent to the model; $c$: the model's tool call; $a_{1:n}$: interpolated end-effector actions; $t$: control-step counter. The success checker evaluates the simulator state after every control step.}
\label{alg:episode}
\begin{algorithmic}[1]
\small
\State $o \gets \textsc{Reset}()$; \; $t \gets 0$
\State $H \gets [\text{system prompt},\ \text{goal},\ o]$
\For{$k = 1, \dots, 80$} \Comment{LLM call budget}
    \State $c \gets \textsc{LLM}(H)$ \Comment{one tool call}
    \If{$c$ is \texttt{done} or \texttt{give\_up}}
        \State \Return \textsc{Failure} \Comment{no action executed}
    \EndIf
    \State $a_{1:n} \gets \textsc{Interpolate}(\text{current pose},\ c.\text{targets})$
    \For{$i = 1, \dots, n$}
        \State $o \gets \textsc{Step}(a_i)$; \; $t \gets t + 1$
        \If{\textsc{SuccessChecker}()}
            \State \Return \textsc{Success}
        \EndIf
        \If{$t = 900$} \Comment{step limit}
            \State \Return \textsc{Failure}
        \EndIf
    \EndFor
    \State $H \gets H + [c,\ \text{feedback}(n),\ o]$ \Comment{incomplete and call finished}
\EndFor
\State \Return \textsc{Failure} \Comment{call budget exhausted}
\end{algorithmic}
\end{algorithm}

\begin{figure}[!p]
\centering
\setcounter{subfigure}{0}

\taskpanel{0.36}{00}{\textbf{Task 0.} Put both the alphabet soup and the tomato sauce in the basket.}\hspace{22pt}
\taskpanel{0.36}{01}{\textbf{Task 1.} Put both the cream cheese box and the butter in the basket.}\\[4pt]

\taskpanel{0.36}{02}{\textbf{Task 2.} Turn on the stove and put the moka pot on it.}\hspace{22pt}
\taskpanel{0.36}{03}{\textbf{Task 3.} Put the black bowl in the bottom drawer of the cabinet and close it.}\\[4pt]

\taskpanel{0.36}{04}{\textbf{Task 4.} Put the white mug on the left plate and the yellow and white mug on the right plate.}\hspace{22pt}
\taskpanel{0.36}{05}{\textbf{Task 5.} Pick up the book and place it in the back compartment of the caddy.}\\[4pt]

\taskpanel{0.36}{06}{\textbf{Task 6.} Put the white mug on the plate and the chocolate pudding to the right of the plate.}\hspace{22pt}
\taskpanel{0.36}{07}{\textbf{Task 7.} Put both the alphabet soup and the cream cheese box in the basket.}\\[4pt]

\taskpanel{0.36}{08}{\textbf{Task 8.} Put both moka pots on the stove.}\hspace{22pt}
\taskpanel{0.36}{09}{\textbf{Task 9.} Put the yellow and white mug in the microwave and close it.}

\caption{\textbf{The ten LIBERO-10 tasks at initial state 0.}
Each panel shows the agentview camera (left) and the wrist camera
(right) at the first decision step.}

\vspace*{\fill}
\label{fig:taskgallery}

\end{figure}

\subsection{Termination and Evaluation}
\label{sec:evaluation}

\paragraph{Per-step success check and the interaction loop.}
After every control step, LIBERO's success checker evaluates the task's goal predicates on the current simulator state. If they are satisfied, the episode ends immediately as a success, even if actions of the current call remain. Otherwise, execution continues with the next action of the same call. When all actions of the call have been executed and the task is still incomplete, the new camera images, robot state, and execution feedback are appended to the history, and the model is queried for the next call. Algorithm~\ref{alg:episode} summarizes one episode.

\paragraph{Termination.}
An episode normally ends in one of five ways:
\begin{enumerate}
    \item the success checker reports success;
    \item the model calls \texttt{done} or \texttt{give\_up};
    \item the model has used all 80 LLM calls, in which case the framework issues \texttt{give\_up} on its behalf;
    \item the episode reaches 900 control steps; this limit is checked after every step, so it can end an episode in the middle of a call; or
    \item the next LLM call would exceed the per-episode spending cap, which guards against runaway API cost.
\end{enumerate}

An episode counts as a success only in case~1, when LIBERO's success checker reports success; episodes that end in any other way count as failures. We report the success rate, the fraction of episodes that end in success.

\section{Results}
\label{sec:results}


\subsection{Tasks and Experimental Setup}

We evaluate frontier AI models on all ten tasks in the LIBERO-10 benchmark~\cite{liu2023libero}, indexed from 0 to 9 and illustrated in Figure~\ref{fig:taskgallery}. The benchmark uses a Franka Emika Panda robotic arm equipped with a parallel-jaw gripper. Each task involves multiple object interactions and requires completing a sequence of manipulation actions. Task success is determined by the corresponding success predicate defined in the original LIBERO benchmark.

Each task provides 50 predefined initial states, of which we select the first ten (indices 0--9) for evaluation, resulting in 100 episodes per model configuration. Before evaluation begins, we execute ten neutral simulation steps with the gripper open to allow the scene to settle. The episode step counter is then reset to zero to ensure that these initialization steps are excluded from the evaluation.



\begin{figure}[t]
\centering
\includegraphics[width=\textwidth]{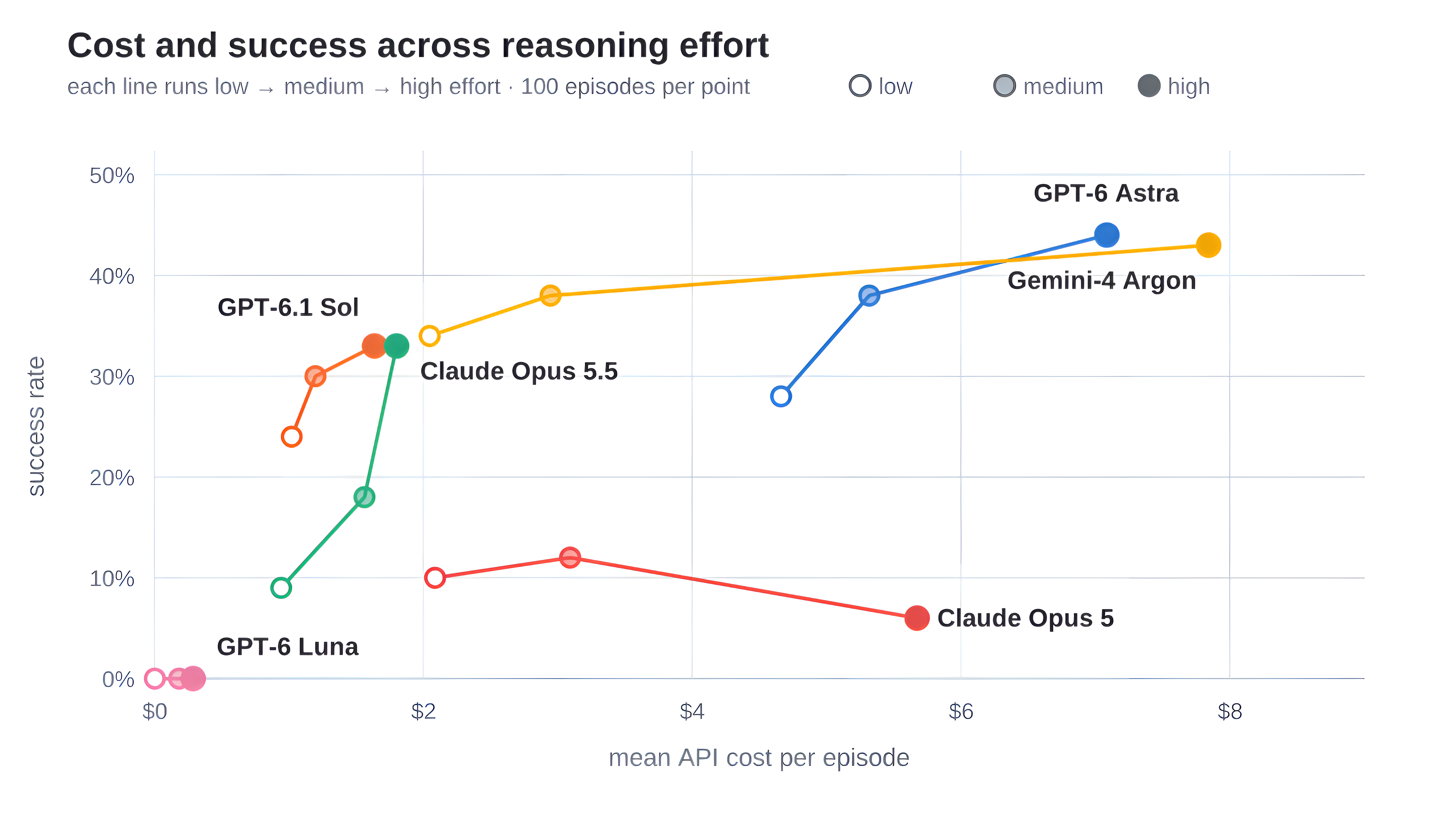}
\caption{\textbf{Task success rate versus average API cost per episode across reasoning effort levels.}
Results are shown for six frontier AI models evaluated on LIBERO-10, with low, medium, and high reasoning effort. Each configuration is evaluated on 100 episodes. Higher reasoning effort generally improves task success but increases inference cost, with substantial differences across models.}
\label{fig:cost}
\end{figure}

\subsection{Impact of Reasoning Effort on Task Success and Inference Cost}
\label{sec:reasoning_effort}

We investigate the impact of reasoning effort on task success and inference cost across frontier AI models on the LIBERO-10 benchmark. For each model, we evaluate three reasoning effort levels (low, medium, and high), with 100 episodes per configuration. Figure~\ref{fig:cost} presents the relationship between mean API cost per episode and task success rate.

The results reveal substantial differences in both performance and cost across models. GPT-6 Astra achieves the highest success rate of 44\% at high reasoning effort, followed closely by Gemini-4 Argon at 43\%. However, both models incur relatively high inference costs, approximately \$7--8 per episode. In contrast, both GPT-6.1 Sol and Claude Opus 5.5 achieve success rate of 33\%, at substantially lower costs of approximately \$1.6--2.0 per episode. GPT-6 Luna achieves zero success across all reasoning effort levels despite its low inference cost. Claude Opus 5 exhibits relatively low success rates, ranging from 6\% to 12\%, even with substantially higher inference costs.

Increasing reasoning effort generally improves task success, although the magnitude of improvement varies considerably across models. GPT-6 Astra improves from 28\% to 44\%, while Gemini-4 Argon improves from 34\% to 43\%, accompanied by a substantial increase in inference cost. GPT-6.1 Sol also benefits from increased reasoning effort, improving from 24\% to 33\%. Claude Opus 5.5 exhibits an improvement from 9\% to 33\%. Interestingly, Claude Opus 5 does not benefit consistently from increased reasoning effort, with its success rate decreasing from 12\% at medium effort to 6\% at high effort.

Overall, these results demonstrate that increased reasoning effort can substantially improve robot manipulation performance, but the benefits are model-dependent and come at additional inference cost. The results also highlight an important trade-off between task success and computational expense. While the most capable reasoning models achieve the highest success rates, more cost-efficient models can provide competitive manipulation performance at a fraction of the cost.

\subsection{Model-Level Performance Analysis}
\label{sec:detailed_results}

Table~\ref{tab:aggregate} summarizes the detailed results across 18 model configurations, including task success, failure outcomes, API costs, and API calls. Across 1,800 evaluation episodes, we achieve an overall success rate of 22.2\%, requiring approximately 99,000 API calls at a total cost of \$4,936.

\begin{table}[h]
\centering
\captionsetup{skip=12pt}
\footnotesize
\setlength{\tabcolsep}{2.5pt}
\begin{tabular}{rlccccccc}
\toprule
\# & Model & Reasoning Effort & Success & Rate & Failure & Limit reached & Cost (\$) & Calls \\
\midrule
1  & GPT-6 Astra     & high   & 44/100 & 44\% &  53 &  3 &   709 & 4{,}936 \\
2  & Gemini-4 Argon  & high   & 43/100 & 43\% &  16 & 41 &   784 & 5{,}376 \\
3  & GPT-6 Astra     & medium & 38/100 & 38\% &  62 &  0 &   532 & 4{,}773 \\
4  & Gemini-4 Argon  & medium & 38/100 & 38\% &  12 & 50 &   295 & 4{,}848 \\
5  & Gemini-4 Argon  & low    & 34/100 & 34\% &  28 & 38 &   205 & 5{,}567 \\
6  & Claude Opus 5.5 & high   & 33/100 & 33\% &  57 & 10 &   180 & 6{,}358 \\
7  & GPT-6.1 Sol     & high   & 33/100 & 33\% &  65 &  2 &   164 & 5{,}376 \\
8  & GPT-6.1 Sol     & medium & 30/100 & 30\% &  69 &  1 &   120 & 5{,}155 \\
9  & GPT-6 Astra     & low    & 28/100 & 28\% &  72 &  0 &   466 & 4{,}833 \\
10 & GPT-6.1 Sol     & low    & 24/100 & 24\% &  75 &  1 &   102 & 5{,}155 \\
11 & Claude Opus 5.5 & medium & 18/100 & 18\% &  73 &  9 &   156 & 6{,}796 \\
12 & Claude Opus 5   & medium & 12/100 & 12\% &  79 &  9 &   309 & 6{,}214 \\
13 & Claude Opus 5   & low    & 10/100 & 10\% &  89 &  1 &   207 & 6{,}152 \\
14 & Claude Opus 5.5 & low    &  9/100 &  9\% &  89 &  2 &  94.2 & 5{,}945 \\
15 & Claude Opus 5   & high   &  6/100 &  6\% &  88 &  6 &   567 & 6{,}442 \\
16 & GPT-6 Luna      & low    &  0/100 &  0\% & 100 &  0 & 0.294 &    852 \\
17 & GPT-6 Luna      & medium &  0/100 &  0\% &  79 & 21 &  18.4 & 6{,}745 \\
18 & GPT-6 Luna      & high   &  0/100 &  0\% &  57 & 43 &  28.7 & 7{,}401 \\
\midrule
\multicolumn{3}{l}{\textbf{Campaign}} & \textbf{400/1800} & \textbf{22.2\%}
& \textbf{1163} & \textbf{237} & \textbf{4{,}936} & \textbf{98{,}924} \\
\bottomrule
\end{tabular}

\caption{\textbf{Evaluation results across 18 model configurations, ranked by success rate.}
Each configuration is evaluated on 100 episodes using the same tasks and initial states. \texttt{Failure} counts episodes terminated by the model through
\texttt{done()} or \texttt{give\_up()} (968) or by reaching the 900-step
horizon (195). \texttt{Limit reached} counts episodes cut off before
the model could finish: by the 80 model-call allowance, the per-episode
spending cap, or a provider output-token limit or API error. Both are counted
as unsuccessful episodes. API costs are estimated based on token usage,
pricing, and cache usage.}

\label{tab:aggregate}
\end{table}

The results also reveal differences in execution behavior across models. Some configurations exhibit a large number of stopped episodes, particularly Gemini-4 Argon and GPT-6 Luna. Interestingly, higher reasoning effort does not necessarily increase the number of API calls. For example, GPT-6 Astra requires approximately 4,770--4,940 calls across all three effort levels, despite substantial differences in inference cost. This suggests that the additional cost of higher reasoning effort primarily comes from more expensive API calls rather than more frequent interactions.


\subsection{Task-Level Performance Analysis}
\label{sec:task_results}

\begin{table}
\centering
\captionsetup{skip=12pt}
\small
\setlength{\tabcolsep}{3pt}
\begin{tabular}{clccccc}
\toprule
ID & Task & Success & Rate & Limit Reached & Cost (\$) & Calls \\
\midrule
1 & cream cheese + butter in basket     & 105/180 & 58\% & 14 & 312 & 8{,}687 \\
0 & alphabet soup + tomato sauce        &  99/180 & 55\% & 14 & 403 & 9{,}485 \\
7 & alphabet soup + cream cheese        &  90/180 & 50\% & 18 & 344 & 9{,}444 \\
2 & turn on stove, moka pot on it       &  59/180 & 33\% & 22 & 496 & 9{,}830 \\
8 & both moka pots on the stove         &  21/180 & 12\% & 29 & 579 & 10{,}373 \\
3 & black bowl in drawer, close it      &  13/180 &  7\% & 24 & 596 & 10{,}752 \\
5 & book into back of caddy             &  12/180 &  7\% & 28 & 493 & 8{,}818 \\
9 & mug in microwave, close it          &   1/180 &  1\% & 40 & 588 & 10{,}310 \\
6 & mug on plate, pudding to its right  &   0/180 &  0\% & 21 & 515 & 10{,}124 \\
4 & two mugs onto left + right plates   &   0/180 &  0\% & 27 & 611 & 11{,}101 \\
\bottomrule 
\end{tabular}
\caption{\textbf{Results by task, over eighteen configurations.} Each task is
attempted 180 times: six models $\times$ three reasoning levels $\times$ ten
initial states. Instructions are abbreviated.}
\label{tab:bytask}
\end{table}

Table~\ref{tab:bytask} presents the evaluation results for each LIBERO-10 task across all 18 model configurations. Task success varies significantly, ranging from 58\% for placing cream cheese and butter in a basket to 0\% for two tasks involving multiple object manipulations. Tasks involving placing objects into a basket achieve relatively high success rates (50--58\%), while tasks requiring more complex interactions, such as opening and closing drawers or microwaves, are considerably more challenging. The two tasks with zero success involve placing multiple objects at specified locations. These results highlight the difficulty of long-horizon manipulation tasks that require precise spatial reasoning and sequential execution. More challenging tasks also tend to incur higher API costs and require more API calls.

\subsection{Per-Model Task Performance}
\label{sec:per_model_results}

Tables~\ref{tab:detail-astra}--\ref{tab:detail-luna} provide detailed task-level results for each model under different reasoning effort levels. GPT-6 Astra demonstrates strong performance across multiple tasks, particularly object placement and stove manipulation, with success rates generally improving as reasoning effort increases. Gemini-4 Argon achieves consistently high success rates on the three basket placement tasks, including perfect success on several configurations. GPT-6.1 Sol exhibits similar strengths at a lower inference cost, although its performance on more complex tasks remains limited.

Claude Opus 5.5 shows substantial improvements with increased reasoning effort, particularly on object placement and stove manipulation tasks, whereas Claude Opus 5 exhibits inconsistent performance and even degrades at high reasoning effort. GPT-6 Luna fails to complete any task across all three reasoning levels. Across models, tasks involving precise spatial arrangements and multi-step interactions remain particularly challenging. These results suggest that reasoning effort improves manipulation capabilities for some models, but does not necessarily overcome the challenges of complex physical interactions.

\begin{table}[H]
\centering
\permodelcap
\caption{\textbf{GPT-6 Astra}}
\label{tab:detail-astra}
\footnotesize
\setlength{\tabcolsep}{2.2pt}
\begin{tabular}{@{}cl *{3}{@{\hspace{10pt}} l@{\hspace{3pt}} c c c}@{}}
\toprule
& & \multicolumn{4}{c}{\textbf{low}} & \multicolumn{4}{c}{\textbf{medium}} & \multicolumn{4}{c}{\textbf{high}} \\
\cmidrule(lr){3-6}\cmidrule(lr){7-10}\cmidrule(lr){11-14}
ID & Task & \multicolumn{2}{c}{Solved\,/10} & \$ & Calls & \multicolumn{2}{c}{Solved\,/10} & \$ & Calls & \multicolumn{2}{c}{Solved\,/10} & \$ & Calls \\
\midrule
0 & soup + tomato sauce & \sbar{7} & 7 & 55.6 & 561 & \sbar{8} & 8 & 42.3 & 446 & \sbar{9} & 9 & 63.5 & 464 \\
1 & cream cheese + butter & \sbar{8} & 8 & 40.2 & 454 & \sbar{10} & 10 & 31.2 & 374 & \sbar{10} & 10 & 35.5 & 357 \\
2 & stove on, moka pot on it & \sbar{2} & 2 & 42.0 & 440 & \sbar{7} & 7 & 50.9 & 468 & \sbar{8} & 8 & 61.9 & 448 \\
3 & bowl in drawer, close it & \sbar{1} & 1 & 52.4 & 496 & \sbar{2} & 2 & 75.0 & 581 & \sbar{2} & 2 & 99.1 & 600 \\
4 & two mugs onto two plates & \sbar{0} & \z & 67.7 & 602 & \sbar{0} & \z & 66.6 & 534 & \sbar{0} & \z & 91.1 & 577 \\
5 & book into back of caddy & \sbar{0} & \z & 25.0 & 322 & \sbar{1} & 1 & 31.3 & 314 & \sbar{0} & \z & 73.2 & 481 \\
6 & mug on plate, pudding R & \sbar{0} & \z & 46.2 & 493 & \sbar{0} & \z & 52.6 & 491 & \sbar{0} & \z & 84.9 & 561 \\
7 & soup + cream cheese & \sbar{8} & 8 & 33.0 & 425 & \sbar{8} & 8 & 41.6 & 442 & \sbar{9} & 9 & 44.6 & 406 \\
8 & both moka pots on stove & \sbar{2} & 2 & 49.8 & 509 & \sbar{2} & 2 & 65.8 & 538 & \sbar{6} & 6 & 60.4 & 462 \\
9 & mug in microwave, close & \sbar{0} & \z & 54.4 & 531 & \sbar{0} & \z & 74.5 & 585 & \sbar{0} & \z & 94.4 & 580 \\
\midrule
\multicolumn{2}{@{}l}{\textbf{All tasks} (of 100)} &  & \textbf{28} & \textbf{466} & \textbf{4{,}833} &  & \textbf{38} & \textbf{532} & \textbf{4{,}773} &  & \textbf{44} & \textbf{709} & \textbf{4{,}936} \\
\bottomrule
\end{tabular}
\end{table}

\begin{table}[!ht]
\centering
\permodelcap
\caption{\textbf{Gemini-4 Argon}}
\label{tab:detail-gemini}
\footnotesize
\setlength{\tabcolsep}{2.2pt}
\begin{tabular}{@{}cl *{3}{@{\hspace{10pt}} l@{\hspace{3pt}} c c c}@{}}
\toprule
& & \multicolumn{4}{c}{\textbf{low}} & \multicolumn{4}{c}{\textbf{medium}} & \multicolumn{4}{c}{\textbf{high}} \\
\cmidrule(lr){3-6}\cmidrule(lr){7-10}\cmidrule(lr){11-14}
ID & Task & \multicolumn{2}{c}{Solved\,/10} & \$ & Calls & \multicolumn{2}{c}{Solved\,/10} & \$ & Calls & \multicolumn{2}{c}{Solved\,/10} & \$ & Calls \\
\midrule
0 & soup + tomato sauce & \sbar{10} & 10 & 8.69 & 373 & \sbar{10} & 10 & 21.8 & 416 & \sbar{10} & 10 & 37.6 & 413 \\
1 & cream cheese + butter & \sbar{8} & 8 & 13.2 & 445 & \sbar{9} & 9 & 17.1 & 387 & \sbar{10} & 10 & 22.2 & 336 \\
2 & stove on, moka pot on it & \sbar{6} & 6 & 17.9 & 480 & \sbar{4} & 4 & 31.1 & 526 & \sbar{6} & 6 & 88.5 & 514 \\
3 & bowl in drawer, close it & \sbar{0} & \z & 22.7 & 675 & \sbar{4} & 4 & 46.7 & 567 & \sbar{2} & 2 & 85.0 & 542 \\
4 & two mugs onto two plates & \sbar{0} & \z & 30.1 & 763 & \sbar{0} & \z & 37.5 & 653 & \sbar{0} & \z & 80.2 & 640 \\
5 & book into back of caddy & \sbar{0} & \z & 18.6 & 501 & \sbar{1} & 1 & 25.2 & 466 & \sbar{2} & 2 & 131 & 709 \\
6 & mug on plate, pudding R & \sbar{0} & \z & 19.7 & 591 & \sbar{0} & \z & 38.8 & 613 & \sbar{0} & \z & 62.9 & 526 \\
7 & soup + cream cheese & \sbar{9} & 9 & 12.0 & 414 & \sbar{10} & 10 & 16.2 & 380 & \sbar{10} & 10 & 27.8 & 370 \\
8 & both moka pots on stove & \sbar{1} & 1 & 29.5 & 605 & \sbar{0} & \z & 45.2 & 593 & \sbar{2} & 2 & 126 & 644 \\
9 & mug in microwave, close & \sbar{0} & \z & 32.2 & 720 & \sbar{0} & \z & 15.0 & 247 & \sbar{1} & 1 & 123 & 682 \\
\midrule
\multicolumn{2}{@{}l}{\textbf{All tasks} (of 100)} &  & \textbf{34} & \textbf{205} & \textbf{5{,}567} &  & \textbf{38} & \textbf{295} & \textbf{4{,}848} &  & \textbf{43} & \textbf{784} & \textbf{5{,}376} \\
\bottomrule
\end{tabular}
\end{table}

\begin{table}[H]
\centering
\permodelcap
\caption{\textbf{GPT-6.1 Sol}}
\label{tab:detail-sol}
\footnotesize
\setlength{\tabcolsep}{2.2pt}
\begin{tabular}{@{}cl *{3}{@{\hspace{10pt}} l@{\hspace{3pt}} c c c}@{}}
\toprule
& & \multicolumn{4}{c}{\textbf{low}} & \multicolumn{4}{c}{\textbf{medium}} & \multicolumn{4}{c}{\textbf{high}} \\
\cmidrule(lr){3-6}\cmidrule(lr){7-10}\cmidrule(lr){11-14}
ID & Task & \multicolumn{2}{c}{Solved\,/10} & \$ & Calls & \multicolumn{2}{c}{Solved\,/10} & \$ & Calls & \multicolumn{2}{c}{Solved\,/10} & \$ & Calls \\
\midrule
0 & soup + tomato sauce & \sbar{8} & 8 & 10.7 & 548 & \sbar{7} & 7 & 11.9 & 522 & \sbar{9} & 9 & 13.3 & 484 \\
1 & cream cheese + butter & \sbar{7} & 7 & 8.84 & 489 & \sbar{5} & 5 & 10.3 & 483 & \sbar{9} & 9 & 8.99 & 397 \\
2 & stove on, moka pot on it & \sbar{0} & \z & 12.6 & 574 & \sbar{4} & 4 & 12.8 & 538 & \sbar{4} & 4 & 18.2 & 551 \\
3 & bowl in drawer, close it & \sbar{1} & 1 & 10.6 & 513 & \sbar{0} & \z & 17.4 & 635 & \sbar{0} & \z & 19.8 & 603 \\
4 & two mugs onto two plates & \sbar{0} & \z & 12.6 & 593 & \sbar{0} & \z & 13.8 & 560 & \sbar{0} & \z & 19.0 & 598 \\
5 & book into back of caddy & \sbar{2} & 2 & 6.86 & 383 & \sbar{4} & 4 & 5.48 & 325 & \sbar{2} & 2 & 14.3 & 480 \\
6 & mug on plate, pudding R & \sbar{0} & \z & 10.5 & 539 & \sbar{0} & \z & 12.6 & 523 & \sbar{0} & \z & 21.4 & 642 \\
7 & soup + cream cheese & \sbar{6} & 6 & 10.8 & 542 & \sbar{8} & 8 & 8.81 & 453 & \sbar{5} & 5 & 11.8 & 470 \\
8 & both moka pots on stove & \sbar{0} & \z & 10.1 & 511 & \sbar{2} & 2 & 13.4 & 551 & \sbar{4} & 4 & 19.9 & 599 \\
9 & mug in microwave, close & \sbar{0} & \z & 8.44 & 463 & \sbar{0} & \z & 13.4 & 565 & \sbar{0} & \z & 17.0 & 552 \\
\midrule
\multicolumn{2}{@{}l}{\textbf{All tasks} (of 100)} &  & \textbf{24} & \textbf{102} & \textbf{5{,}155} &  & \textbf{30} & \textbf{120} & \textbf{5{,}155} &  & \textbf{33} & \textbf{164} & \textbf{5{,}376} \\
\bottomrule
\end{tabular}
\end{table}

\begin{table}[H]
\centering
\permodelcap
\caption{\textbf{Claude Opus 5.5}}
\label{tab:detail-opus55}
\footnotesize
\setlength{\tabcolsep}{2.2pt}
\begin{tabular}{@{}cl *{3}{@{\hspace{10pt}} l@{\hspace{3pt}} c c c}@{}}
\toprule
& & \multicolumn{4}{c}{\textbf{low}} & \multicolumn{4}{c}{\textbf{medium}} & \multicolumn{4}{c}{\textbf{high}} \\
\cmidrule(lr){3-6}\cmidrule(lr){7-10}\cmidrule(lr){11-14}
ID & Task & \multicolumn{2}{c}{Solved\,/10} & \$ & Calls & \multicolumn{2}{c}{Solved\,/10} & \$ & Calls & \multicolumn{2}{c}{Solved\,/10} & \$ & Calls \\
\midrule
0 & soup + tomato sauce & \sbar{1} & 1 & 9.87 & 635 & \sbar{4} & 4 & 13.3 & 639 & \sbar{8} & 8 & 12.1 & 508 \\
1 & cream cheese + butter & \sbar{2} & 2 & 8.31 & 544 & \sbar{7} & 7 & 10.6 & 544 & \sbar{9} & 9 & 12.1 & 533 \\
2 & stove on, moka pot on it & \sbar{3} & 3 & 9.31 & 612 & \sbar{3} & 3 & 14.7 & 669 & \sbar{8} & 8 & 17.8 & 638 \\
3 & bowl in drawer, close it & \sbar{0} & \z & 9.52 & 584 & \sbar{1} & 1 & 18.6 & 739 & \sbar{0} & \z & 21.7 & 712 \\
4 & two mugs onto two plates & \sbar{0} & \z & 10.7 & 632 & \sbar{0} & \z & 18.7 & 748 & \sbar{0} & \z & 21.3 & 679 \\
5 & book into back of caddy & \sbar{0} & \z & 9.07 & 544 & \sbar{0} & \z & 16.9 & 718 & \sbar{0} & \z & 20.7 & 694 \\
6 & mug on plate, pudding R & \sbar{0} & \z & 7.98 & 498 & \sbar{0} & \z & 15.6 & 665 & \sbar{0} & \z & 18.5 & 604 \\
7 & soup + cream cheese & \sbar{3} & 3 & 9.53 & 617 & \sbar{3} & 3 & 14.1 & 669 & \sbar{8} & 8 & 16.7 & 675 \\
8 & both moka pots on stove & \sbar{0} & \z & 10.3 & 669 & \sbar{0} & \z & 16.2 & 726 & \sbar{0} & \z & 17.9 & 691 \\
9 & mug in microwave, close & \sbar{0} & \z & 9.63 & 610 & \sbar{0} & \z & 17.7 & 679 & \sbar{0} & \z & 21.3 & 624 \\
\midrule
\multicolumn{2}{@{}l}{\textbf{All tasks} (of 100)} &  & \textbf{9} & \textbf{94.2} & \textbf{5{,}945} &  & \textbf{18} & \textbf{156} & \textbf{6{,}796} &  & \textbf{33} & \textbf{180} & \textbf{6{,}358} \\
\bottomrule
\end{tabular}
\end{table}

\begin{table}[H]
\centering
\permodelcap
\caption{\textbf{Claude Opus 5}}
\label{tab:detail-opus5}
\footnotesize
\setlength{\tabcolsep}{2.2pt}
\begin{tabular}{@{}cl *{3}{@{\hspace{10pt}} l@{\hspace{3pt}} c c c}@{}}
\toprule
& & \multicolumn{4}{c}{\textbf{low}} & \multicolumn{4}{c}{\textbf{medium}} & \multicolumn{4}{c}{\textbf{high}} \\
\cmidrule(lr){3-6}\cmidrule(lr){7-10}\cmidrule(lr){11-14}
ID & Task & \multicolumn{2}{c}{Solved\,/10} & \$ & Calls & \multicolumn{2}{c}{Solved\,/10} & \$ & Calls & \multicolumn{2}{c}{Solved\,/10} & \$ & Calls \\
\midrule
0 & soup + tomato sauce & \sbar{4} & 4 & 18.6 & 574 & \sbar{3} & 3 & 29.7 & 635 & \sbar{1} & 1 & 48.7 & 665 \\
1 & cream cheese + butter & \sbar{5} & 5 & 14.8 & 525 & \sbar{4} & 4 & 28.7 & 637 & \sbar{2} & 2 & 44.7 & 667 \\
2 & stove on, moka pot on it & \sbar{0} & \z & 22.1 & 630 & \sbar{3} & 3 & 33.3 & 610 & \sbar{1} & 1 & 58.4 & 669 \\
3 & bowl in drawer, close it & \sbar{0} & \z & 21.0 & 615 & \sbar{0} & \z & 30.0 & 651 & \sbar{0} & \z & 61.8 & 708 \\
4 & two mugs onto two plates & \sbar{0} & \z & 23.9 & 659 & \sbar{0} & \z & 38.4 & 689 & \sbar{0} & \z & 74.4 & 666 \\
5 & book into back of caddy & \sbar{0} & \z & 18.7 & 563 & \sbar{0} & \z & 24.7 & 448 & \sbar{0} & \z & 68.7 & 575 \\
6 & mug on plate, pudding R & \sbar{0} & \z & 22.7 & 609 & \sbar{0} & \z & 34.5 & 624 & \sbar{0} & \z & 60.7 & 581 \\
7 & soup + cream cheese & \sbar{0} & \z & 22.4 & 700 & \sbar{2} & 2 & 28.2 & 633 & \sbar{1} & 1 & 41.0 & 600 \\
8 & both moka pots on stove & \sbar{1} & 1 & 22.9 & 636 & \sbar{0} & \z & 32.6 & 628 & \sbar{1} & 1 & 55.0 & 690 \\
9 & mug in microwave, close & \sbar{0} & \z & 19.5 & 641 & \sbar{0} & \z & 29.1 & 659 & \sbar{0} & \z & 54.0 & 621 \\
\midrule
\multicolumn{2}{@{}l}{\textbf{All tasks} (of 100)} &  & \textbf{10} & \textbf{207} & \textbf{6{,}152} &  & \textbf{12} & \textbf{309} & \textbf{6{,}214} &  & \textbf{6} & \textbf{567} & \textbf{6{,}442} \\
\bottomrule
\end{tabular}
\end{table}

\begin{table}[H]
\centering
\permodelcap
\caption{\textbf{GPT-6 Luna}}
\label{tab:detail-luna}
\footnotesize
\setlength{\tabcolsep}{2.2pt}
\begin{tabular}{@{}cl *{3}{@{\hspace{10pt}} l@{\hspace{3pt}} c c c}@{}}
\toprule
& & \multicolumn{4}{c}{\textbf{low}} & \multicolumn{4}{c}{\textbf{medium}} & \multicolumn{4}{c}{\textbf{high}} \\
\cmidrule(lr){3-6}\cmidrule(lr){7-10}\cmidrule(lr){11-14}
ID & Task & \multicolumn{2}{c}{Solved\,/10} & \$ & Calls & \multicolumn{2}{c}{Solved\,/10} & \$ & Calls & \multicolumn{2}{c}{Solved\,/10} & \$ & Calls \\
\midrule
0 & soup + tomato sauce & \sbar{0} & \z & 0.035 & 97 & \sbar{0} & \z & 2.08 & 721 & \sbar{0} & \z & 3.16 & 784 \\
1 & cream cheese + butter & \sbar{0} & \z & 0.022 & 67 & \sbar{0} & \z & 1.94 & 714 & \sbar{0} & \z & 2.82 & 734 \\
2 & stove on, moka pot on it & \sbar{0} & \z & 0.030 & 85 & \sbar{0} & \z & 1.76 & 632 & \sbar{0} & \z & 3.15 & 746 \\
3 & bowl in drawer, close it & \sbar{0} & \z & 0.029 & 84 & \sbar{0} & \z & 2.00 & 719 & \sbar{0} & \z & 2.65 & 728 \\
4 & two mugs onto two plates & \sbar{0} & \z & 0.035 & 97 & \sbar{0} & \z & 1.89 & 682 & \sbar{0} & \z & 2.95 & 729 \\
5 & book into back of caddy & \sbar{0} & \z & 0.024 & 75 & \sbar{0} & \z & 1.52 & 593 & \sbar{0} & \z & 2.38 & 627 \\
6 & mug on plate, pudding R & \sbar{0} & \z & 0.026 & 81 & \sbar{0} & \z & 2.14 & 720 & \sbar{0} & \z & 3.05 & 763 \\
7 & soup + cream cheese & \sbar{0} & \z & 0.040 & 104 & \sbar{0} & \z & 2.12 & 764 & \sbar{0} & \z & 2.95 & 780 \\
8 & both moka pots on stove & \sbar{0} & \z & 0.027 & 82 & \sbar{0} & \z & 1.15 & 517 & \sbar{0} & \z & 2.67 & 722 \\
9 & mug in microwave, close & \sbar{0} & \z & 0.026 & 80 & \sbar{0} & \z & 1.78 & 683 & \sbar{0} & \z & 2.88 & 788 \\
\midrule
\multicolumn{2}{@{}l}{\textbf{All tasks} (of 100)} &  & \textbf{0} & \textbf{0.294} & \textbf{852} &  & \textbf{0} & \textbf{18.4} & \textbf{6{,}745} &  & \textbf{0} & \textbf{28.7} & \textbf{7{,}401} \\
\bottomrule
\end{tabular}
\end{table}

\subsection{Failure Analysis Through Model Hindsight}
\label{sec:res:failures}

\paragraph{Failure and termination behaviors.}
Of the 1{,}800 evaluation episodes, 1{,}400 fail to satisfy the task success
predicate. These end in three ways. The model stops the episode itself in 968
cases, by calling \texttt{give\_up} (681) or \texttt{done} (287). A further 195
are stopped by the 900-step simulator horizon, and the remaining 237 are cut
off by the 80-call allowance or the spending cap. The first two groups form the
\emph{Failure} column of Table~\ref{tab:aggregate} and the third its
\emph{Limit reached} column.

The two ways a model stops itself are informative in opposite directions. A
\texttt{give\_up} call means the model has judged the task unreachable; a
\texttt{done} call means it believes the task finished while the success
predicate is still unsatisfied, so its count measures how often a model
mistakes an incomplete task for a complete one.

\paragraph{Learning from model hindsight.}
Our framework requires each model to provide a \texttt{hindsight} field when calling \texttt{done} or \texttt{give\_up}, describing what it wishes it had known at the beginning of the episode. These self-reported observations provide valuable insights into the challenges encountered during manipulation, including spatial understanding, robot control, grasping, and task-specific interactions. For example, GPT-6.1 Sol at low reasoning effort provides the following hindsight after a failed episode:

\begin{quote}\small
``At low heights, forward reach requests were heavily clamped and caused substantial wrist tilt instead of reaching the target. Check actual pose rather than requested pose; fully closed finger positions repeatedly revealed missed grasps. The wrist camera grasp point is near the top of the image, not its center. The stove control is behind the burner, and straight pushes did not reliably actuate it.''
\end{quote}

This example illustrates how the model identifies limitations in motion execution, visual grounding, grasp verification, and task-specific interactions. Although these observations are generated by the models themselves and may not always accurately explain the causes of failure, they provide useful hypotheses for understanding unsuccessful manipulation attempts.

\paragraph{Common challenges identified from hindsight.}
We analyze the hindsight notes to identify recurring challenges across models. One prominent issue is the discrepancy between commanded and executed robot motions. Specifically, 16\% of Gemini-4 Argon’s failure notes, 42\% of GPT-6 Astra’s, 47\% of GPT-6.1 Sol’s, 56\% of GPT-6 Luna’s, 88\% of Claude Opus 5.5’s, and 91\% of Claude Opus 5’s mention that \texttt{move\_to} commands do not fully reach their targets. Models frequently attribute this behavior to motion clamping and suggest repeating commands or checking the actual robot state.

We further verify this observation by comparing commanded end-effector targets with the positions achieved at the next interaction step. The achieved displacement averages approximately 64--72\% of the commanded displacement across these models, with 21--36\% of motions covering less than half the requested distance. This confirms that incomplete motion execution is a common challenge across models, although the frequency with which models identify it in hindsight varies substantially.

\paragraph{Implications for improving robot manipulation.}
The hindsight analysis suggests that successful manipulation requires more than high-level reasoning and action planning. Models must also understand the robot's embodiment, interpret visual observations accurately, and account for discrepancies between commanded and executed actions. Providing more explicit information about robot kinematics, camera geometry, and controller behavior may help reduce these failures. More broadly, model-generated hindsight offers a promising source of feedback for improving prompts, action interfaces, and closed-loop manipulation strategies in future work.

\section{Conclusion}
\label{sec:conclusion}

In this report, we presented a systematic evaluation of six frontier AI models on the LIBERO-10 robot manipulation benchmark using a unified closed-loop manipulation framework. Across 18 model configurations and 1,800 evaluation episodes, the best-performing model, GPT-6 Astra, achieved a success rate of 44\%, demonstrating the potential of general-purpose AI models to perform robot manipulation without task-specific training or fine-tuning.

Our experiments reveal three key findings. First, increasing reasoning effort generally improves manipulation success, although the benefits vary across models and come at additional inference cost. Second, task difficulty has a substantial impact on performance. While frontier AI models perform reasonably well on object placement tasks, they struggle with precise spatial arrangements and complex multi-step interactions. Third, inference costs vary significantly across models, highlighting the importance of considering both task success and computational efficiency.

Overall, our results demonstrate both the promise and current limitations of frontier AI models for robot manipulation. Despite their strong reasoning capabilities, reliably translating visual observations and high-level reasoning into precise physical actions remains challenging. Future research should explore improved spatial grounding, action representations, and integration with low-level robot control and learned manipulation policies. We release our evaluation framework and experimental results to facilitate reproducible benchmarking and future research on reasoning-based robot manipulation.

\bibliography{experiment_references}

@misc{arena2026leaderboard,
  author={{Arena AI}},
  title={{Arena} Leaderboard},
  year={2026},
  howpublished={\url{https://arena.ai/leaderboard}},
  note={Accessed: 2026-10-15}
}

@article{zhang2026unexpected,
  title={An Unexpected Robot Policy: Early Evaluations of GPT-6 Astra on RoboDojo and Beyond},
  author={Zhang, Wenbo and Wang, Kaixuan and Ouyang, Yutao and Huang, Xiaoyu and Li, Liyang and Su, Kailun and Jin, Weiyang and Chai, Wenhao and Liang, Haotian and Dou, Zhiyang and others},
  journal={arXiv preprint arXiv:2609.24170},
  year={2026}
}

@misc{yang2026astranomy,
  title        = {Astranomy: The study of {GPT-6 Astra} as a native robot policy},
  author       = {Yang, Jenai Xuning and Khetan, Ashna and Zhang, Shun},
  year         = {2026},
  month        = {October},
  organization = {Seattle Robotics Lab (SRL), NVIDIA},
  howpublished = {\url{https://research.nvidia.com/labs/srl/projects/astranomy/}},
  note         = {Blog post},
}

@misc{su2026astra,
  title        = {{GPT 6 Astra} as an Embodied Policy},
  author       = {Su, Jiayi and Zheng, Yixin and Yan, Mi and Yi, Li and Zhang, Zhizheng and Wang, He},
  year         = {2026},
  howpublished = {Technical report and code},
  url          = {https://github.com/anonymous-report-421/eval-of-gpt-6-astra-as-policy}
}

@article{team2026systematically,
  title={Systematically Exploring the Capabilities of GPT-6 Astra as Embodied Policies},
  author={Team, Galbot and Chen, Xuchuan and Cheng, Xiaoqian and Deng, Yu and Ding, Lihe and Dong, Shaocong and Gao, Xiangjun and Jia, Haozhe and Li, Zekai and Li, Zhoujian and others},
  journal={arXiv preprint arXiv:2609.38537},
  year={2026}
}

@misc{liu2023libero,
  title         = {{LIBERO}: Benchmarking Knowledge Transfer for Lifelong
                   Robot Learning},
  author        = {Bo Liu and Yifeng Zhu and Chongkai Gao and Yihao Feng and
                   Qiang Liu and Yuke Zhu and Peter Stone},
  year          = {2023},
  eprint        = {2306.03310},
  archivePrefix = {arXiv},
  primaryClass  = {cs.AI},
  url           = {https://arxiv.org/abs/2306.03310}
}

@article{bu2025univla,
  title={Univla: Learning to act anywhere with task-centric latent actions},
  author={Bu, Qingwen and Yang, Yanting and Cai, Jisong and Gao, Shenyuan and Ren, Guanghui and Yao, Maoqing and Luo, Ping and Li, Hongyang},
  journal={arXiv preprint arXiv:2505.06111},
  year={2025}
}

@article{kim2024openvla,
  title={Openvla: An open-source vision-language-action model},
  author={Kim, Moo Jin and Pertsch, Karl and Karamcheti, Siddharth and Xiao, Ted and Balakrishna, Ashwin and Nair, Suraj and Rafailov, Rafael and Foster, Ethan and Lam, Grace and Sanketi, Pannag and others},
  journal={arXiv preprint arXiv:2406.09246},
  year={2024}
}

@article{intelligence2025pi_,
  title={$\pi_{0.5}$: a Vision-Language-Action Model with Open-World Generalization},
  author={Intelligence, Physical and Black, Kevin and Brown, Noah and Darpinian, James and Dhabalia, Karan and Driess, Danny and Esmail, Adnan and Equi, Michael and Finn, Chelsea and Fusai, Niccolo and others},
  journal={arXiv preprint arXiv:2504.16054},
  year={2025}
}

@inproceedings{zheng2025universal,
  title={Universal actions for enhanced embodied foundation models},
  author={Zheng, Jinliang and Li, Jianxiong and Liu, Dongxiu and Zheng, Yinan and Wang, Zhihao and Ou, Zhonghong and Liu, Yu and Liu, Jingjing and Zhang, Ya-Qin and Zhan, Xianyuan},
  booktitle={2025 IEEE/CVF Conference on Computer Vision and Pattern Recognition (CVPR)},
  pages={22508--22519},
  year={2025},
  organization={IEEE}
}

@article{xiao2026enpire,
  title={ENPIRE: Agentic Robot Policy Self-Improvement in the Real World},
  author={Xiao, Wenli and Xie, Jia and Zhang, Tonghe and Lin, Haotian and Fu, Letian and Xue, Haoru and Lu, Jalen and Yang, Yi and Dai, Cunxi and Wang, Zi and others},
  journal={arXiv preprint arXiv:2606.19980},
  year={2026}
}

@article{team2024octo,
  title={Octo: An open-source generalist robot policy},
  author={Team, Octo Model and Ghosh, Dibya and Walke, Homer and Pertsch, Karl and Black, Kevin and Mees, Oier and Dasari, Sudeep and Hejna, Joey and Kreiman, Tobias and Xu, Charles and others},
  journal={arXiv preprint arXiv:2405.12213},
  year={2024}
}

@article{lu2026aspire,
  title={ASPIRE: Agentic/Skills Discovery for Robotics},
  author={Lu, Runyu and Wu, Yubo and Kou, Ethan and Fu, Letian and Xiao, Wenli and Mandlekar, Ajay and Xu, Yinzhen and Shi, Guanya and Goldberg, Ken and Chen, Ang and others},
  journal={arXiv preprint arXiv:2607.00272},
  year={2026}
}

@article{chen2026gap,
  title={GaP: A Graph-as-Policy Multi-Agent Self-Learning Harness For Variational Automation Tasks},
  author={Chen, Kaiyuan and Xie, Shuangyu and Fu, Letian and Yu, Justin and Pacini, William and Bajamahal, Sandeep and Kim, Hudson and Drake, Jaimyn and Kim, Daehwa and Xue, Haoru and others},
  journal={arXiv preprint arXiv:2607.05369},
  year={2026}
}

@misc{ora2026,
  title        = {{ORA}: Open Robot-Use Agents},
  author       = {Yang, Yuanbo and Chen, Siyi and Gao, Jun and Zhang, Hao},
  year         = {2026},
  month        = oct,
  howpublished = {Hao AI Lab blog},
  url          = {https://haoailab.com/blogs/ora/}
}

@software{inspect-robots,
  author  = {Robocurve},
  title   = {Inspect Robots: The open-source evaluation framework for physical AI},
  year    = {2026},
  url     = {https://github.com/robocurve/inspect-robots},
  license = {MIT}
}

\end{document}